\pdfoutput=1

\documentclass[11pt]{article}

\usepackage[]{acl}
\usepackage[T1]{fontenc}
\usepackage[utf8]{inputenc}
\usepackage{times}
\usepackage{latexsym}
\usepackage{amsmath}
\usepackage{amssymb}
\usepackage{amsfonts}
\usepackage{amsthm}
\usepackage{adjustbox}
\usepackage{booktabs,makecell}
\usepackage{bbm}
\usepackage{bm}
\usepackage{textcomp}
\usepackage{mathtools}
\usepackage{xspace}
\usepackage{graphicx}
\usepackage{makecell}
\usepackage{comment}
\usepackage{scalerel}
\usepackage{marvosym}
\usepackage{bclogo}
\usepackage{enumitem}

\usepackage{empheq}
\usepackage[most]{tcolorbox}

\usepackage{algorithm}
\usepackage[noend]{algpseudocode}
\algnewcommand{\parState}[1]{\State%
    \parbox[t]{\dimexpr\linewidth-\algmargin}{\strut\hangindent=\algorithmicindent \hangafter=1 #1\strut}}

\algrenewcommand\algorithmicindent{1.0em}%

\theoremstyle{definition}

\usepackage{cleveref}

\definecolor{darkblue}{RGB}{0,0,160}

\usepackage{times}
\usepackage{latexsym}

\usepackage[T1]{fontenc}

\usepackage[utf8]{inputenc}

\usepackage{microtype}

\newlength\savewidth\newcommand\shline{\noalign{\global\savewidth\arrayrulewidth
  \global\arrayrulewidth 1pt}\hline\noalign{\global\arrayrulewidth\savewidth}}
\newcommand{\tablestyle}[2]{\setlength{\tabcolsep}{#1}\renewcommand{\arraystretch}{#2}\centering\footnotesize}

\title{Prompt, Condition, and Generate: Classification of Unsupported Claims with In-Context Learning }

\author{Peter Ebert Christensen \\
  \\
  \texttt{pec@di.ku.dk} \\\And
  Srishti Yadav \\
  University of Copenhagen \& Pioneer Centre for AI \\
  \texttt{srya@di.ku.dk} \\\And
  Serge Belongie \\
  \\
  \texttt{s.belongie@di.ku.dk} }

\begin{document}
\maketitle
\begin{abstract}
Unsupported and unfalsifiable claims we encounter in our daily lives can influence our view of the world. Characterizing, summarizing, and -- more generally -- making sense of such claims, however, can be challenging. In this work, we focus on fine-grained debate topics and formulate a new task of distilling, from such claims, a countable set of narratives. 
We present a crowdsourced dataset of 12 controversial topics, comprising more than 120k arguments, claims, and comments from heterogeneous sources, each annotated with a narrative label. We further investigate how large language models (LLMs) can be used to synthesise claims using In-Context Learning. We find that generated claims with supported evidence can be used to improve the performance of narrative classification models and, additionally, that the same model can infer the stance and aspect using a few training examples. Such a model can be useful in applications which rely on narratives , e.g. fact-checking. 
\end{abstract}

\section{Introduction}
Online platforms have revolutionized the landscape of public discourse, facilitating extensive debates across a wide range of topics. However, these online discussions often suffer from a lack of coherent and concise arguments. Despite this inherent challenge, it is possible to discern particular motions \cite{levy-etal-2014-context}, opinions \cite{li-etal-2020-exploring-role}, human values \cite{kiesel-etal-2022-identifying}, and narratives \cite{christensen2022PAPYER} within the seemingly disorganized discourse. The ability to identify narratives in online debates is paramount for fact-checking and argument mining, as it enables the evaluation of unsupported claims and their validity.

In our methodology, narratives are differentiated from arguments and claims by incorporating additional attributes: topics, stances, aspects, and evidence.  The  \textit{topic} refers to the subject under discussion, such as the ethical aspects of cloning humans for reproductive purposes. The \textit{stance} represents the viewpoint taken on the topic, for example, a negative stance indicating that cloning for reproductive purposes is considered unethical and unacceptable. Within the narrative, the \textit{aspect} focuses on a specific perspective, providing a more nuanced understanding of the topic. For instance, the aspect within the context of cloning could be the creation of cloned embryos solely for research purposes, which delves into a particular subtopic. These attributes aid in identifying arguments in non-argumentative sources \cite{stab-etal-2018-cross}.

Evidence plays a crucial role in assessing the credibility of a statement. When supported by evidence, a statement gains strength and credibility, classifying it as an argument \cite{hansen-hershcovich-2022-dataset}. For instance, in the text ``Cloning humans for reproductive purposes is unethical and unacceptable, but creating cloned embryos solely for research -- which involves destroying them anyway -- is downright criminal,'' the presence of evidence highlighting the destruction of embryos strengthens the argument. Conversely, without evidence, a statement such as ``Cloning humans for reproductive purposes is unethical'' would be categorized as a claim, lacking the necessary substantiation to be considered an argument. Additional support is required to validate a claim as an argument. With these definitions, we can briefly differentiate between narrative, arguments and claims as: 
\vspace{-2mm}
\begin{enumerate}
\setlength{\itemsep}{-3pt}
\item Narrative: Concise expression of an individual's perspective on a specific topic.
\item Claim: Statement or proposition without supporting evidence.
\item Argument: Claim supported by evidence and reasoning. Aim to justify a specific stance on a topic
\vspace{-2mm}
\end{enumerate}

As seen above, \textit{claims} can lack or have insufficient evidence and be unverifiable or unfalsifiable for purposes of fact-checking in real world scenarios \cite{glockner2022missing} and are hence often not suitable for fact-checking pipelines and thus discarded \cite{augenstein2021explainable}.
Instead of discarding the claims and arguments, we propose that one should instead identify the individual unsupported claim or \textit{narrative}, e.g., ``human cloning is wrong''. We call this task Narrative Prediction which forms the basis of this paper. We use the word ``Prediction'' as an umbrella term, as it can be viewed as either classification (due to the small set of unsupported claims that reflect different viewpoints in the debate) or alternatively a generation task.
In addition to fact-checking the existing literature on claim generation using large language models (LLMs) lacks attention to the relationship between narratives extracted from online debate portals \cite{christensen2022PAPYER} and argumentative texts \cite{habernal-gurevych-2016-argument}, as well as the effective modeling of narratives by LLMs. This work addresses these gaps by formalizing narratives in online debates and understanding their elements: topics, aspects, stance, and evidence. Additionally, a curated dataset of 120k tweets, with around 40 narratives per topic, is introduced to train and evaluate narrative prediction techniques for fact-checking systems.
Furthermore, we propose a method to enhance narrative prediction by generating synthetic tweets through argumentative attributes such as stance and aspects using few-shot In Context Learning (ICL) as illustrated this in Fig.~\ref{fig:PCGdiagram}. The task of narrative prediction simply corresponds to only the right hand side of the figure using no generated candidates where we fine-tune the LLM on tweets. In summary, the contributions of this paper are:

\begin{figure*}[ht]
	\centering
    \includegraphics[width=0.9\textwidth]{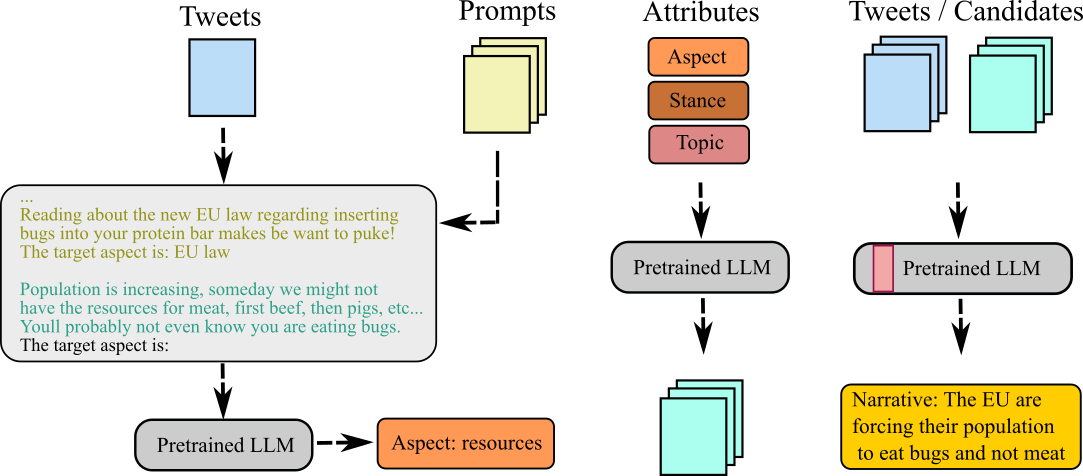}
	\caption{Prompt, Condition, and Generate: A framework to enhance narrative prediction by synthetizing tweets. We first \textit{prompt} a LLM for the stance and aspects of a new tweet using ICL with some examples, we then \textit{condition} the LLM on these attributes to synthesize tweets. Lastly, we fine-tune a LLM on all tweets to \textit{generate} narratives. }
	\label{example_revisions}
    \vspace{-5mm}
    \label{fig:PCGdiagram}
\end{figure*}

\begin{enumerate}
\vspace{-2mm}
\setlength{\itemsep}{-3pt}
\item
\emph{A specific definition} for narratives, along with an analysis of how this differs from arguments, claims, and motions.
\item
\emph{A new dataset and task}, consisting of online comments and tweets labelled for narrative prediction.
\item
\emph{A narrative prediction approach} that maps all the tweets from a fine-grained debate into a list of narratives using a LLM.
\item
\emph{A computational approach} that generates synthetic arguments/claims with a specified aspect and stance. 

\end{enumerate}

\section{Related Works}
\label{sec:relatedwork}



Corpora of textual claims considering various controversial topics have often been used in the study of rhetoric and argumentation, including  summarization \cite{Stammbach}, optimization \cite{skitalinskaya-etal-2022-optimization}, identifying human values \cite{kiesel-etal-2022-identifying}, robustness of arguments \cite{sofi-etal-2022-robustness}, controllable text generation \cite{schiller-etal-2021-aspect}, stance detection \cite{stab-etal-2018-cross}, and studying what constitutes an argument \cite{Trautmann_Daxenberger_Stab_Schütze_Gurevych_2020}.
Prior work on claim and argument summarization has been beneficial in different tasks and domains. In early works, summarization was used for explainable fact-checking \cite{Stammbach, mishra-etal-2020-generating} and has recently been used to denoise tweets \cite{bhatnagar-etal-2022-harnessing}. However, abstractive summarization techniques for real-world tweets are still underdeveloped compared to traditional text summarization methods. Given the effectiveness of prompt-based methods in tasks like abstractive summarization and binary classification \cite{chung2022scaling, sanh2022multitask}, we propose exploring these methods to enhance the text generation of arguments, particularly within the fine-grained topic debates. While fine-grained approaches have been explored in argument mining \cite{hansen-hershcovich-2022-dataset, Trautmann_Daxenberger_Stab_Schütze_Gurevych_2020, schiller-etal-2021-aspect}, they often address broader controversial topics (``minimum wage.'') rather than narrow debates (``crypto currencies as a fiat currency,''). 
Similarly other works that classify if a claim is mentioned in a text \cite{mirkin-etal-2018-listening} studies motions which are an action that should be taken (as can be seen with the example ``we should introduce goal line technology'').
Other lines of works focus on scaling up by detecting ``generic'' claims frequent across topics \cite{orbach-etal-2019-dataset} or mining candidate claims from corpora \cite{lavee-etal-2019-towards}. In comparison
we are envisioning our work to be applicable for unfalsifiable or unverifiable claims coming from short noisy tweets rather than a high quality curated database (iDebate) which contain minutes long speeches.

In our work we create a new dataset, focusing on narrow debate topics, by relying on an argument mining annotation scheme based on \citet{hansen-hershcovich-2022-dataset}, consisting of various categories of claims and arguments found in online debates. Where \citet{hansen-hershcovich-2022-dataset} compare arguments in terms of categories (normative or factual arguments), we propose and study the new task of predicting controversial narratives from tweets. Perhaps most similar to our work is \citet{christensen2022PAPYER}, which proposed a human-in-the-loop-based model to cluster unfalsifiable claims using crowdsourced triplets similarities. 
\section{Task and Data}

This section introduces our definition of a narrative, and a proposed task, and presents the data used for development and evaluation.

\subsection{Narrative Definition}
As mentioned in the introduction, we define the term \textit{narrative} as a concise statement lacking supporting evidence, which can originate from an unfalsifiable or unverifiable claim. Additionally, narratives can include arguments supported by evidence types such as anecdotal, expert, or normative sources as defined in ~\cite{hansen-hershcovich-2022-dataset}, instead of empirical studies. The objective of our methodology is to identify a small set of unsupported claims that reflect diverse viewpoints in a debate and require attention from fact-checkers.


After defining the term narrative we can now focus on a theoretical underpinning of this paper which is a proposed relationship between number of narratives and the scope of the fine-grained debate, that we call the \textit{parrot hypothesis}.


\paragraph{The Parrot Hypothesis} 
In a given social media debate, the thoughts and opinions contributed by commentators resolve to a finite set of distinct narratives. While users could, in principle, state their views in a concise, distilled manner, they often prefer to write embellished variants or personal takes that require reading between the lines. 


At its core, the parrot\footnote{We use ``parrot'' in the sense of ``parroting talking points,'' except that we don't assume the commentators are necessarily being fed talking points without their knowledge.} hypothesis seeks to propose a concept to manage the variations of statements in a debate. By grouping statements into a finite set of narratives related to common topics, the hypothesis narrows the scope of the debate and transforms it into a classification problem. Narratives, representing individual unsupported claims or viewpoints, play a crucial role in capturing diverse perspectives and supporting fact-checking efforts. Despite the potential for infinite arguments, a limited number of distinct claims tend to emerge in online debates, backed by the majority of users \cite{boltuzic-snajder-2015-identifying}. Our hypothesis is that a narrow enough topic will emit such behaviour from users. Incorporating the parrot hypothesis and identifying narratives could enable a more systematic analysis to improve understanding of narratives and facilitating fact-checking.

\subsection{Narrative Prediction}

We approach the problem of narrative prediction on social media, specifically focusing on tweets. 

\paragraph{Task} Given a single tweet $t$, a statement by a participant in a debate, and a set of possible narratives $\mathcal{N}$, rewrite $t$ into a narrative $n\in \mathcal{N}$ such that:

\begin{itemize}[noitemsep]
    \item the narrative is written as an unsupported claim,
    \item only one narrative $n$ can be selected for each tweet from $\mathcal{N}$, and 
    \item $n$ preserves the meaning of $t$ as much as possible.
\end{itemize}

The set of possible narratives, denoted as $\mathcal{N}$, is sourced from domain experts. Although a tweet may implicitly or explicitly contain multiple narratives, our aim is to identify and assign only one explicitly stated narrative for each tweet.

By addressing narrative prediction in this manner, we strive to transform tweets into coherent and explicit unsupported claims, contributing to a deeper understanding and analysis of the content within the context of social media discourse.

\subsection{Annotation scheme}

To collect relevant data, we use an annotation scheme comprising a fine-grained topic, a sentence, and a narrative (unfalsifiable and unverfiable claim). Additionally, we explore the augmentation of an existing dataset, following an alternative annotation scheme \cite{schiller-etal-2021-aspect}, to incorporate attributes such as stance (polarity of the argument) and aspect (subtopics or viewpoints) and use it for the generation of synthetic tweets. 
Though there exist narratives that are claims with supporting evidence (can be anecdotal, factual or normative which are found in \cite{hansen-hershcovich-2022-dataset}), the type of evidence is not considered for annotation. 
The details of this will be explained in the following section.

\subsection{Dataset creation}
We present two datasets: Twitter-Narratives-9 (TN9) and an augmented version of the UKP-Corpus-Aug dataset. UKP-Corpus-Aug, which is the augmented UKP-Corpus dataset includes stance, aspect, and narrative annotations for three randomly selected topics from the original UKP-Corpus \cite{schiller-etal-2021-aspect, stab-etal-2018-cross}. On the other hand, TN9 consists of narrative annotations for 9 carefully selected controversial topics. Table \ref{tab:datasets} provides an overview of the datasets, including key statistics and a comparison between them. Additionally, Table \ref{tab:dataset-example} presents examples from TN9.

\begin{table}[h]
\label{tab:datasets}
\vspace{-1mm}
\begin{center}
\resizebox{\columnwidth}{!}{%
\begin{tabular}{lcc}
\toprule 
 & \bf UKP-Corpus-Aug & \bf TN9  \\
\midrule
\bf Annotations  & Aspect/Stance/Narrative & Narrative  \\
\bf Tweets (train/test) & 30k/1.9k & 90k/5.4k\\

\bf Topics & Abortion, Cloning,   & AGI, Attractiveness,  \\
& Nuclear Energy  & Alternative Meat,   \\
&    & Corporate culture, \\
&    & Crypto, Baby Formula,  \\
&    & Influencer, Transport,  \\
&    & Mental health  \\
\bf Source (Sentence) & Reddit & Twitter
\\
\bf Source (Labels) & mTurk & mTurk \\
\bottomrule
\end{tabular}%
}
\end{center}
\caption{Summary of the datasets. (The original UKP-Corpus consists of only Stance and Aspects, we provide narrative annotations)}
\vspace{-3mm}
\label{tab:datasets}
\end{table}

\begin{table}[h]
    \centering
    \adjustbox{width=\columnwidth}{
        \setlength\tabcolsep{2.5pt}
        \begin{tabular}{lll}
        \toprule 
        \bf Topic & \bf Sentence &  \bf Narrative \\
        \addlinespace[0.1cm]
        Crypto & 
        you are promoting crypto which is  &  Influencers are scamming \\
         & a scam helping thieves and criminals   & their fans using crypto \\
         & you are also full of plastic parts  & \\
         & and fillers profitable for the  & \\
         &  pharmaceutical and cosmetic industry & \\
        \midrule 
        Formula &
         My congressman here voted NO on , & People are reselling baby   \\
         &  lowering gas prices, NO on the baby  & formula to other countries \\
         & formula bill, NO on contraception (?!!),  & for higher prices  \\
         &  and NO on   other helpful bills.  It is   & \\
         &  unbecoming to complain about economic & \\
         &   hardship and then contribute to it. & \\
        \midrule 
        AGI & And on the other side, AGI will be  & AI will not replace \\
        & the single greatest technology to  &  humans but augment them \\
        & alleviate human suffering in all of history. & \\

        \bottomrule
        \end{tabular}
    }
    \caption{Example sentences and annotated narratives.} \label{tab:dataset-example}
\vspace{-5mm}
\end{table}

\subsubsection{Scraping}
We start with scraping relevant data from Twitter. First a series of searches is executed combining different keywords and sentences/phrases, highlighting different statements in a topic. We search for 40 different keywords per topic from year 2016-2022 and search for as many fields (e.g., images, links, and other metadata) as possible using the Twitter API. The specific keywords used for each topic can be found in Appendix B.

\subsubsection{Filtering and Data Cleaning}

\begin{figure}[h]
    \vspace{-3mm}
    \centering
    \includegraphics[width=\columnwidth]{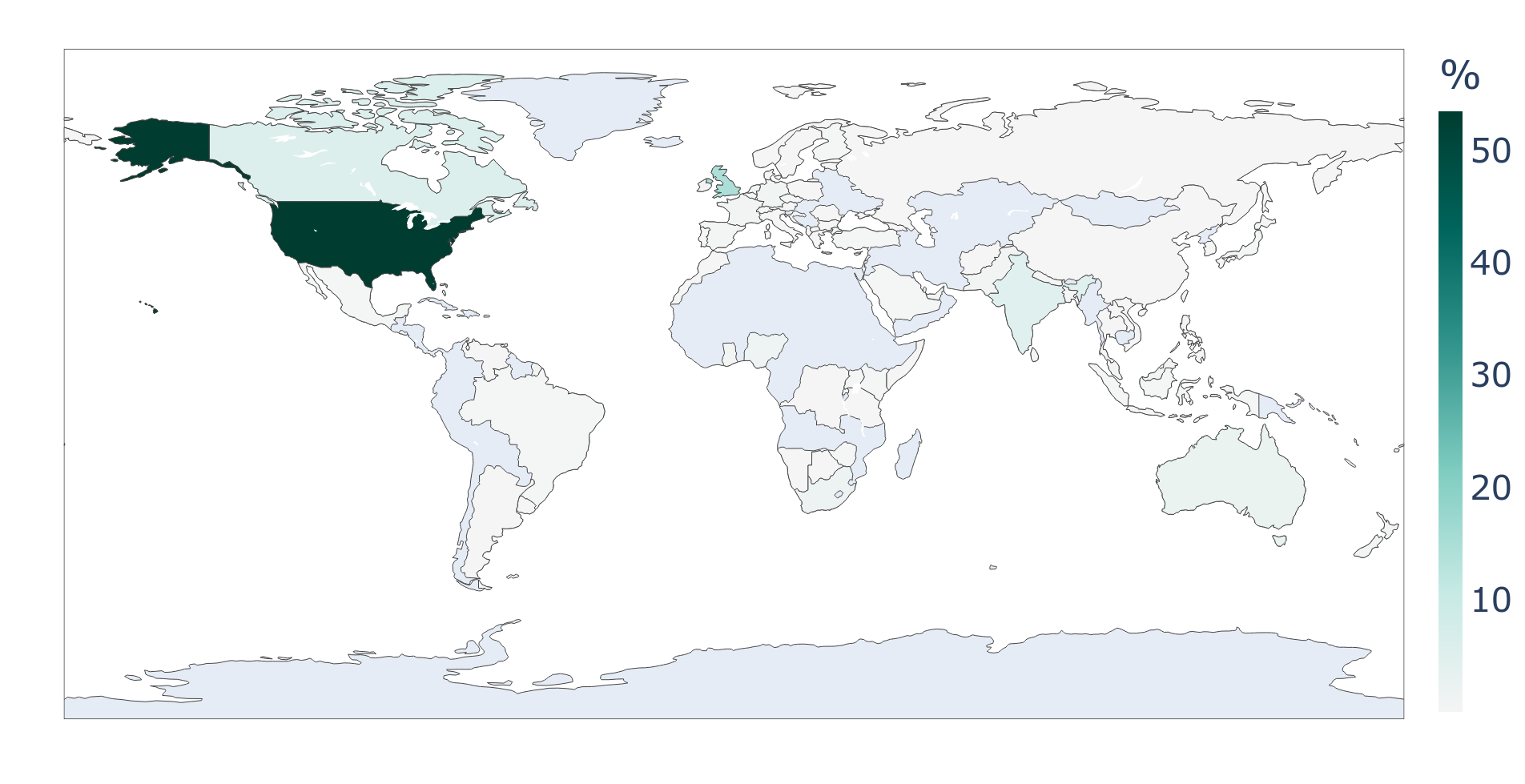}
    \vspace{-8mm}
    \caption{Visualization of the percentages of the number of tweets per country. Like \cite{huang2019largescale} only 2\% of all tweets had available geotags and the tweets are found to be predominantly from the US, where the userbase is numerically the largest.}
    \label{fig:world-tweet-percentages}
\vspace{-6mm}
\end{figure}

To ensure that we are working with claims, we perform filtering steps. First, we remove duplicates but maintain identical sentences with different hashtags after removing retweets, quote tweets, links and videos, as well as mentions of users, token and media mentions. Second, we replace unreadable hexadecimal representations of unicode characters with their respective character, and encode the text with ascii characters. This results in 98,187 English tweets in total, around 11k tweets for each topic. The geographic distribution of the tweets as shown in Figure \ref{fig:world-tweet-percentages}.

\subsubsection{Dataset Annotation}
Annotation of narratives is conducted using Amazon Mechanical Turk in 2 rounds. 
In round 1, we design a pretest to ensure that the workers know the difference between an argument, claim and evidence using the dataset from \cite{hansen-hershcovich-2022-dataset}.
Given a tweet the annotators classify whether the tweet is a claim, argument or neither. Furthermore they also classify the evidence type given an argument \cite{hansen-hershcovich-2022-dataset}. Learning to distinguish between these claims will help them in determining the narrative of the tweet. After passing at least 4 out of 5 questions the workers could begin annotating our 120k tweets.
Using lists of around 40 narratives per topic compiled by domain experts we proceed to round 2. Each task consists of 1 tweet from 1 topic and annotators are asked to pick 1 out of $\sim$ 40 different narratives this tweet follows (if any). 
Furthermore, by using the definition of argument as in \citet{hansen-hershcovich-2022-dataset} we can decompose it into 1) a claim and 2) its evidence, and use their schema to categories the type of evidence an argument may have. We thus additionally ask if the tweet is a claim or an argument with an evidence type that is not a study. A study here refers to ``Results of a quantitative analysis of data, given as numbers, or as conclusions.'' That is statements that are cited, or easy to verify numbers should they appear in another argument. \cite{hansen-hershcovich-2022-dataset}
The pay is 18\$/hr.
Detailed instructions can be found in Appendix D.

\section{Method}\label{sec:method}

\paragraph{Problem statement:} Our goal is to create a model that output an estimate of the true narrative $n$ given tweet $t$ from debate $d$. We do so by 
\begin{itemize}
    \item Investigating if identifying the narrative of a tweet is best suited as a text2text approach or a classification approach.
    \item Creating a data augmentation step using different kinds of ICL to help the best fine tuning procedure.
\end{itemize}

\paragraph{Narrative prediction approach}
Given observed data $\{t_i,n_i\}_{i=1}^N$, we could parameterize a model as $\Tilde{n}=f_{\phi}(t)$, where $f$ is a pretrained LLM with parameters $\phi$ is the model that we finetune. This is illustrated as step 3 in \cref{fig:PCGdiagram}. We note that our prediction $\Tilde{n}$ in this case would be free-form text, a text2text approach.
As the last step in \cref{fig:PCGdiagram}. illustrates a finetuning procedure we could alternatively parameterise a model as $\Tilde{n}=g(h_{\theta}(f_{\omega}(t)))$, where $g$ is a lookup function that maps class $c_i$ to narrative $n_i$ for debate $d$, $h$ is a multi-class classifier with parameters $\theta$, and $f$ the model from before, but only using the encoder with parameters $\omega$ and the classification head $h$.The prediction $\Tilde{n}$ is now a class, a multi-class classification approach.
The lookup function $g$ enables us to take a predicted class and look up the actual narrative in the list written in appendix F. But doing this substitution we can calculate a Rouge score between the target narrative and the predicted one. However, during training we opt for optimizing the crossentropy loss on the narrative classes.

\paragraph{Prompt, Condition and Generate}
In addition to only finetuning $f$, which is a LLM, we argue for using new methodologies focusing on ICL to exploit the capabilities of the LLM further and validate their performance on our new dataset. 
As such we let us inspire by prior work \citet{schiller-etal-2021-aspect} on generating synthetic arguments (candidates) using aspects and a stance, that we call Prompt, Condition and Generate (PCG) and add candidates to our finetuning procedure as illustrated in step 3 in \cref{fig:PCGdiagram}.
In contrast to their work our setup requires no training and can be done using a few examples using ICL as illustrated in \cref{fig:PCGdiagram}. 
That is as a first step we annotate a few handmade examples with topics names, a binary stance and a snippet of the example tweet which forms the aspect and then we \textit{prompt} the frozen model to output the stance and aspect of a novel tweet $t$. 
Then we \textit{condition} the same model anew on its predicted stance and aspect to generate a candidate, by asking it to write a tweet knowing only about the debate topic, a stance and the aspect.
To complete the creation of synthetic data we copy the original narrative $n$ from $t$ to form the candidate. This data can be used for additional fine-tuning of the text generation model that \textit{generate} narratives, as shown in \cref{fig:PCGdiagram}.

\paragraph{ICL Methods}
In the context of using LLM for ICL, a simple but effective approach called few-shot learning is to provide several examples of a task in the same prompt with the given input \cite{brown2020language}.
Additionally one can also first generate an explanation as to why certain outputs are favourable before generating the final answer, this is called Chain-of-Thought (CoT) \cite{wei2023chainofthought}. Furthermore one should be careful with the selected examples for ICL. As noted in \citet{zhao2021calibrate}, standard ICL can be biased towards the training examples and the order of their occurrence. To mitigate this effect one can estimate the bias towards each answer by feeding in a test input that is content-free, e.g., ``N/A'' and ``''. In practice on can fit an affine transformation to ``calibrate'' (Cal) the model's output probabilities to cause a uniform prediction for ``N/A''.
We will investigate these methods in our PCG setup using different numbers of shots for aspect and stance prediction.

\section{Experiments}

In this section we investigate the performance of our finetuning approaches, including synthetic tweet generation for performance enhancement and the subtasks of stance and aspect prediction using different ICL techniques. We predict narratives on 7548 test cases (629 per topic).

\subsection{Setup}

\paragraph{Classification:} 
As described in Section 4.1, our classification model (\textit{SFT\_{head}}) consists of an encoder $f_{\omega}$, being a T0 encoder \cite{sanh2022multitask} model and a multi-class classifier $h$, which is a single MLP that project the hidden layer down to the number of narratives present in one topic. We only finetune $h$ using  using the crossentropy loss. Finally using $g$ we can report the Rouge-L F1 score as we convert the predicted class into the written narrative and comparing it with ground truth.

\paragraph{Text generation:} In contrast to the classification model $f_{\phi}$ is the full T0 model. We add new parameters and make a parameter efficient fine-tuning setup known as LoRA \cite{liu2022fewshot} on the T0 model. LoRA incorporates two low-rank matrices that are added to each parameter matrix in T0. We measure the Rougle-L F1 score between the generated narrative $\hat{n}$ and the ground truth narrative.

\paragraph{Prompt, Condition and Generate:} To enhance the above mentioned setups we generate synthetic tweets. We do this we first infer the stance and aspect of a new tweet by insert up to 4 such examples into the prompt, and then second simply prompt our frozen model to write a tweet with the predicted stance containing the sentence from the aspect and about the same topic as the new tweet, this is shown in Figure \ref{fig:PCGdiagram}.

To test how well the model can predict aspects and stances we focus on stance and aspect data from the UKP-Corpus on 3 randomly select topics, as these sentences are annotated with a stance and aspects. We compare 3 methods: standard ICL, CoT \cite{wei2023chainofthought} and Cal \cite{zhao2021calibrate}, with a fully supervised BERT span predictors as a baseline. We train 3 BERT$_{BASE}$  baselines with $_{only}$ indicating it is \textit{only} trained on this topic using 10k examples. The second $_{remain}$ uses 60k total examples training on the 5 \textit{remaining}  out of distribution topics from \cite{stab-etal-2018-cross} ( i.e. excluding abortion, cloning \& nuclear energy) before fine-tuning to a new topic and finally $_{all}$  trained on all 8 topics (80k examples) from \cite{schiller-etal-2021-aspect}.The ICL setups use 4 examples to predict the attributes, though we experiment with including fewer examples for aspect prediction (Figure \ref{fig:scaling-codex} ).
We do not use verbalizers for ICL but restrict the possible decoding output only to the words considered in the sentence for aspect prediction or ``for'' or ``against'' in case of stance prediction. 

In addition to generating candidates with our original model $f$, we test the generality of generating candidates of these attributes by conditioning other LLMs on them. These include T5-Flan-3B \cite{chung2022scaling}, BLOOM-175B \cite{workshop2023bloom} and CTRLUKP \cite{schiller-etal-2021-aspect} for the UKP-Corpus.  When finetuning using candidates and tweets, we compare them using several metrics like precision oriented BLEU \cite{papineni-etal-2002-bleu}, recall oriented Rouge-L \cite{lin-2004-rouge}, METEOR \cite{banerjee-lavie-2005-meteor}, and finally chrF \cite{popovic-2015-chrf}.
To automatically quantify to what extent a candidate contains the meaning of the original claim, we compute their semantic similarity in each case using the BERT-score\cite{bert-score}.
Additionally we conduct a human evaluation of the generated candidates to ensure readability for humans. For each generative model and topic we select 10 candidates and acquire 2 independent crowdworkers via MTurk at 18\$/hour.
The annotators scored all candidates on four quality metrics: (1) argument quality (2) persuasiveness, (3) meaning preservation and (4) fluency. 
We follow \citet{schiller-etal-2021-aspect} for assessing the argument quality, \citet{habernal-gurevych-2016-argument} for persuasiveness and \citet{skitalinskaya-etal-2022-optimization} for quality using these Likert scales:
Argument Quality. 1 (much worse than original) - 5 (notably improved),  Persuasiveness.  1 (generated text less persuasive than original) - 3 (generated text is more persuasive), Fluency. 1 (major errors) - 3 (fluent) and Meaning Preservation. 1 (entirely different) - 5 (identical).
Lastly we report the inter-annotator agreement \cite{cohen_kappa} and krippendorff's alpha \cite{krippendorff04} between 2 annotators.

\paragraph{Stance prediction} To do stance prediction, we classify a tweet as either ``for'' or ``against'' a particular topic. For ICL methods we output the most likely word and convert it to 0 or 1 to compare it with the binary class output for the baseline model. 
We use binary cross entropy loss to compare the predicted stance with the true label.
\paragraph{Aspect prediction} Here we aim to identify the correct span of text within a tweet. The span is represented using the beginning-inside-outside (BIO) tags format \cite{ramshaw1995text}. Here the initial word within the span is given the label ``B'' for beginning, the following words within the span is given the label ``I'' for inside, and finally any other word is given the label ``O'' for outside, making it a ternary classification task for the baselines.
We sample multiple completions using beam search and report the average micro F1 and accuracy for both stance and aspect prediction.

\subsection{Results}

\paragraph{Classification} 10 epochs of fine-tuning LM head results in a 38.54 Rouge-L F1 score for the UKP corpus and 38.72 Rouge-L F1  score for TN9.

\paragraph{Text generation}
Fine-tuning the LoRA weights results in a 39.32 Rouge-L F1 score for the UKP corpus and 39.49 Rouge-L F1 score for TN9, similar to other summarization tasks \cite{zhang-etal-2022-summn}. 
Inspecting the results of this models for a couple of outputs is shown in Table \ref{tab:example-predictions}. Analysing the second example we see a more concisely written narrative than the target, this lowers the resulting Rouge F1 score due to its shorter common sub-sequence.
\begin{table}[h]
    \centering
    \adjustbox{width=\columnwidth}{
        \setlength\tabcolsep{2.5pt}
        \begin{tabular}{l|l|l}
        \toprule 
        \bf Tweet $t$ & \bf Model Prediction $t_n$ &  \bf Target Narrative $n$ \\
        \midrule 
        Animals are not ingredients! &eating meat is murder &Eating meat is murder  \\
        \midrule 
        Yall find hypermasculinity &&\\
        resulting in insecurities &&\\
        about the lack of a better &Hypermasculinity is &Hypermasculinity in and\\  
         body attractive? Lmaaoo &problematic & of itself is the problem \\


        \bottomrule
        \end{tabular}
    }
    \caption{Sentences with predicted and target narrative.} \label{tab:example-predictions}
    \vspace{-5mm}
\end{table}

\paragraph{Prompt condition generate}
Given prior results we proceed with the best setup, the text generation setup from before and Table \ref{tab:cls_init} shows the average Rouge-L F1 micro accuracy using additional candidate examples generated by T0-3B, T5-flan-3B model \cite{chung2022scaling}, an API call to BLOOM-176B \cite{workshop2023bloom} and the CTRL generative model from \citet{schiller-etal-2021-aspect} respectively. 
Using 629 candidates we get a 4 percentage point increase from the 39.49 Rouge-L F1 for TN9 from before, highlighting the strength of our approach. 

\begin{table}[h]
\vspace{-1mm}
\begin{center}
\begin{tabular}{lcc}
\toprule 
Setups & \bf UKP & \bf TN9  \\
\midrule
SFT\_{T0}& 43.56   & 43.89   \\
SFT\_{T5F}& 44.12  & 44.34  \\
SFT\_{BLOOM}& 42.64   & 42.97  \\
SFT\_{CTRL}& 43.34  &  --   \\
\bottomrule
\end{tabular}%
\end{center}
\vspace{-2mm}

\caption{Summary of Rouge-L F1 scores using text2text supervised fine-tuning on the original dataset as well as candidates generated with different models. 
} \label{tab:cls_init}
\vspace{-4mm}
\end{table}

Table \ref{tab:exp1-auto} shows the quantitative metrics of our candidates. The relatively low BLEU (6.5) and ROUGE-L (9.2) indicate that revisions take place, however due to the high BERT-score (90.5) the meaning is largely preserved. Also the METEOR and Rouge-L scores are similar to \citet{schiller-etal-2021-aspect} indicating similar generative behaviour. 
TN9 lower scores indicate that the model has difficulty generating similar sentences to the original tweet using a predicted stance and aspect. 
\begin{table}[h]
    \renewcommand{\arraystretch}{0.95}
    \setlength{\tabcolsep}{2pt}
    \centering
    \small
    \begin{tabular}{@{}l@{\hspace*{-0.75em}}rrrcr}
        \toprule
        \bf Approach & 
        \multicolumn{1}{l}{\bf BLEU} &
        \multicolumn{1}{r}{\bf RouL} &
        \multicolumn{1}{l}{\bf Meteor} &
        \multicolumn{1}{r}{\bf BERT-score} &
        \multicolumn{1}{c}{\bf chrf}  \\

        \midrule
        
\bf UKP-Corpus \\
 \, CTRLUKP & 8.3 &  12.1 & 16.4 & 83.7 & 23.1 \\  
 \, BLOOM &  6.5 &  13.6  & 16.2 & 84.8 & \bf31.1     \\

  \, T5-flan    		& 10.8     & \bf20.6 		& 16.4 	& \bf90.5    & 25.1      	\\
  
  \,	T0    		& \bf 13.6      & 20.3	&\bf 16.7  & 90.2  & 25.2        	\\[0.5ex]

\bf TN9 \\
        		
 \, BLOOM         		& 7.94	    &  9.2 		& 9.7	& 82.1    & 23.8     \\

  \, T5-flan    		& 11.2      & 13.7 		& 9.5 & 87.4    & 18.7      	\\
  
  \,	T0    		& 12.3      & 13.1 		& 9.2  & 87.8  & 18.9     	\\[0.5ex]

        	 
\bottomrule
\end{tabular}
\caption{Automatic evaluation: Average performance of each model on 629 test cases per topic}
\label{tab:exp1-auto}
\vspace{-4mm}
\end{table}

\begin{figure*}[h]
    \centering 
    \includegraphics[width=0.9\textwidth]{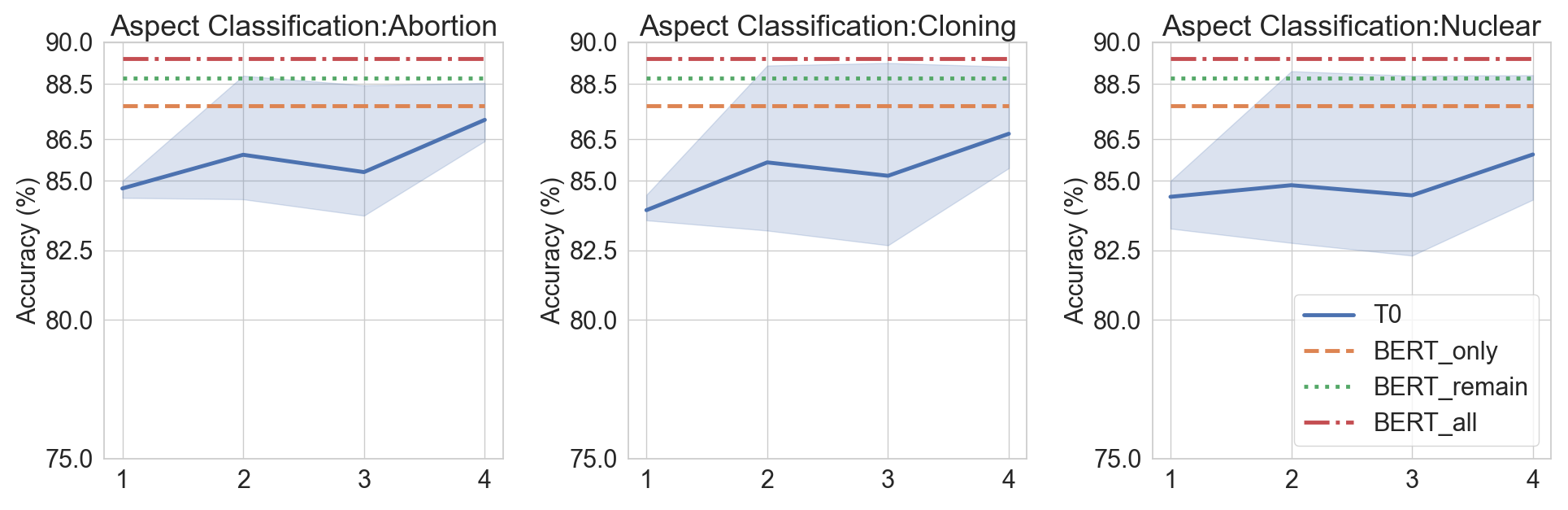}
    \caption{Average Aspect accuracy of few-shot ICL (\texttt{T0-3B}) on the Abortion, Cloning and the Nuclear Energy topic in the UKP dataset using random subsets of $k'= 1\ldots4$ examples. We display the best performances of the best fine-tuned BERT$_{BASE}$ baselines, the tags $_{only}$,  $_{remain}$ and $_{all}$ indicate the same setup from table \ref{tab:results-per-debate-method-stance}.}
\label{fig:scaling-codex}
\vspace{-3mm}

\end{figure*}

Table \ref{tab:exp1-human} show generally low Krippendorff’s alpha agreement of 0.24 on average, which are common in subjective tasks \cite{wachsmuth-etal-2017-computational}. The inter-annotator agreement \cite{cohen_kappa} varies from model and attribute but is on average .25, which can be interpreted as “fair” agreement \cite{landis}. 
Table \ref{tab:exp1-human} shows that human annotators find text generated by T0, having a higher persuasiveness (2.6) and having similar meaning to the source text (4.5) than the other methods. However, candidates from BLOOM and CTRL-UKP have a higher argument quality (3.5 vs. 3.6 and 4.2) and are more fluently written. 
Table \ref{tab:interannot} shows T0 being preferred for generating meaningful and persuasive texts. This is important as we will use the data in a fine-tuning setup.
\begin{table}[h]
    \renewcommand{\arraystretch}{0.95}
    \setlength{\tabcolsep}{2pt}
    \centering
    \small
    \begin{tabular}{@{}l@{\hspace*{-0.75em}}cccc}
        \toprule
        \bf Model &
        \bf Persuasiveness  & 
        \multicolumn{1}{l}{\bf Fluency} &
        \multicolumn{1}{l}{\bf Argument} &
        \multicolumn{1}{l}{\bf Meaning}  \\

        \midrule
        
\bf UKP-Corpus \\
 \, CTRLUKP          &  2.1 	& 2.3 		& 3.6  		&  3.4  \\
 \, BLOOM         &   	1.9	&  \textbf{2.8} 		& \textbf{4.2}		& 4.1 	\\
 \, T5-flan        &  	2.2	&  1.8 		&  3.2 		& 3  \\
 \, T0  &	\textbf{2.6}	&  \textbf{2.8} 	& 3.5  		&  \textbf{4.5}\\

\bf TN9 \\
 \, BLOOM          & 2	         &  \textbf{2.7}  & 3.4           & \textbf{3.5}  	\\
 \, T5-flan        & \textbf{2.4} &  2.3           &  \textbf{3.6} & 3.3  \\
 \, T0             &	\textbf{2.4} &  2.5           & 3.4           &  \textbf{3.5}	\\
\bottomrule
\end{tabular}
\caption{Human evaluation: Average scores on 10 candidates per topic using different models.} 
\label{tab:exp1-human}
\vspace{-4mm}
\end{table}
\begin{table}[h]
    \renewcommand{\arraystretch}{0.95}
    \setlength{\tabcolsep}{2pt}
    \centering
    \small
    \begin{tabular}{@{}l@{\hspace*{-0.75em}}cccc}
        \toprule
        \bf Model &
        \bf Persuasiveness  & 
        \multicolumn{1}{l}{\bf Fluency} &
        \multicolumn{1}{l}{\bf Argument} &
        \multicolumn{1}{l}{\bf Meaning}  \\

        \midrule
        
 \, CTRLUKP          &  0.2/0.3 	& 0.2/0.3 		& 0.2/0.2  		&  0.4/0.4  \\
 \, BLOOM         &   	-0.1/-0.1 &  0.4/0.4 		& 0.1/0.2		    & 0.3/0.1 	\\
 \, T5-flan   &  	0.1/0.1	& 0.3/0.3 		&  0.1/0.4 		& 0.5/0.4  \\
 \, T0  &	0.3/0.3 &  0.4/0.3  	& 0.2/0.3  		&  0.5/0.3 \\
\bottomrule
\end{tabular}
\caption{Annotator agreement (Cohens kappa and krippendorffs alpha) using 2 annotators across all topics. }
\vspace{-5mm}
\label{tab:interannot}
\end{table}

\paragraph{Stance prediction}

Table \ref{tab:results-per-debate-method-stance} shows our methods outperform the baselines on at least two topics in UKP-Corpus with Cloning topic being an exception. We believe this is because the distribution of stances in this topic makes it highly polarized. While imperfect it suffices to generate candidates.

\begin{table}[h]
    \centering 
    \resizebox{\columnwidth}{!}{%
    \begin{tabular}{|c|c|c|c|}
    \hline
    \textbf{Method} & \textbf{Abor. (F1 / Acc)} & \textbf{Clon. (F1 / Acc)} & \textbf{Nucl. (F1 /  Acc)} \\ \hline
    only                 & 50.1 / 53.1                         & 75.5 / 75.8                        & 37.1 / 58.9                          \\ \hline
    ICL                  & 54.4 / 53.8                         & 59.3 / 54.9                        & 54.9 / 52.9                       \\ \hline
    CoT                  & 55.7 / 54.7                         & 62.4 / 56.7                        & 57.4 / 53.6                        \\ \hline
    Cal                  & 57.3 / 55.6                         & 60.6 / 55.9                       & 58.7 / 54.1                        \\ \hline
    remain               & 36.1 / 56.4                           & 35.6 / 55.3                          & 37.1 / 58.9                          \\ \hline
    all                  & 52.6 / 55.6                         & 77.1 / 77.6                        & 37.1 / 58.9                          \\ \hline
    \end{tabular}
    }
\caption{Average micro F1 and accuracy for stance prediction using binary classification (for=1,against=0).}  \label{tab:results-per-debate-method-stance}
\vspace{-6mm}
\end{table}

\paragraph{Aspect prediction}
Table \ref{tab:results-per-debate-method-aspect} shows that our method perform worse than our best baseline trained on 80k examples, but perform at a similar level to the official results reported in Table 3 in  \cite{schiller-etal-2021-aspect}. Additionally our baseline increases performance on the number of topics it has been trained on.
Figure \ref{fig:scaling-codex} visualises the performance of T0 few-shot prediction given $k \le 4$ examples and baseline models.
\texttt{T0-3B} using 4 tweets is competitive to baselines trained on 10k+ tweets, despite the variance of the predictions being rather large, which reflect that using the samples is not always beneficial to the model. We proceed with the ICL setup for generating candidates despite this.

\begin{table}[h]
    \centering 
    \resizebox{\columnwidth}{!}{%
    \begin{tabular}{|c|c|c|c|}
    \hline
    \textbf{Method} & \textbf{Abor. (F1 / Acc)} & \textbf{Clon. (F1 / Acc)} & \textbf{Nucl. (F1 /  Acc)} \\ \hline
    only                 & 68.5 / 87.7                         & 71.8 / 88.9                        & 73.1 / 89.9                          \\ \hline
    ICL                  & 66.9 / 87.1                         & 66.5 / 86.6                       & 66.1 / 86.3                        \\ \hline
    CoT                  & 67.2 / 87.3                         & 67.7 / 87.7                        & 68.8 / 88.3                        \\ \hline
    Cal                  & 68.2 / 87.8                         & 68.5 / 88.2                      & 68.4 / 88.1                        \\ \hline
    remain               & 71.6 / 88.7                           & 74.9 / 90.5                      & 75.5 / 91                          \\ \hline
    all                  & 72.9 / 89.4                         & 75.2 / 90.9                        & 76.6 / 91.5                         \\ \hline
    \end{tabular}
    }
\caption{Average micro F1 and accuracy for aspect prediction using BIO tags. } 
\label{tab:results-per-debate-method-aspect}
\vspace{-6mm}

\end{table}

\section{Conclusion}

In this paper we introduced a new definition of narratives and how to model these in fine-grained debates with large language models. Our approach is based on parameter efficient fine-tuning using controlled text generation using attributes predicted using a handful of examples. 
 We show that claims generated using our approach are genuine and sensible in general. We fine-tune of model on our own dataset and the augmented UKP-corpus and outperform baseline approaches.
 In future work, we seek to examine multiple completions and ensembles similar to \cite{liévin2023large} which enables to include examples of up to 100 examples for ICL, to reduce variance and outperform single-sample CoT methods using larger models (GPT-4, ChatGPT, LLama). Moreover, our approach considers each topic independently using a LLM but could be made to consider all simultaneously. 

\section*{Acknowledgements}
PEC, SY and SB was supported by the Pioneer Centre for AI, DNRF grant number P1.
\section{Limitations}

\paragraph{Scaling to multiple topics}


For  our  approach, the prediction of narratives is topic specific and the number of models scales linearly with the with the topics. This is primarily because both the baseline method using a LM head cannot predict new classes and for the text2text approach it is theoretically possible to simply use one model, though initial experiments suggested a model per topic worked better. 
Instead of directly predicting the narratives, one could instead have ranked the list of narratives given a tweet.
This gives us contextual information about the narratives, since they are written in text and not just as a class and provides a number of benefits including having one model for all topics but also new topics. Additionally it could also provide temporal evaluations by adding new emerging narratives to the list. 

\paragraph{Scaling to more narratives} The current approach requires a domain expert to writing down the particular narratives from the fine grained debate and does not model that there is a countable number of narratives within a specific domain.
Finding the particular narratives is bottlenecked by knowing enough about the particular topic. Moreover, since it takes time to gather enough information about the different topics it makes it difficult to scale up to larger numbers of taxons. 

Future work can explore automatic generation of the narratives given a list of tweets, and condense this list iteratively, and 
patch templates e.g., using pre-trained language
models.



\paragraph{Directly modelling the initial argumentative text}
Finally, the approach we develop can operate on text that is claims or argument discourse units, but has no way of distinguishing between these or nonarguments.
This precludes the model from being able to only predict a narrative if the text is indeed from the fine grained debate and can be tricked into providing narratives which the text doesn't follow.




\bibliography{custom}
\bibliographystyle{acl_natbib}

\clearpage

\appendix

\section{Implementation Details}
Here we describe the implementation details for fine-tuning the 3B T0 model for narrative prediction, in addition to using different ICL strategies for stance and aspect prediction.
For all downstream tasks,
we use the same AdamW optimiser with linear learning rate decay and weight decay.
Finetuning details such as number of epochs and learning rate is reported in \autoref{tab:impl_finetune} 

\paragraph{Aspect prediction}
For our aspect prediction models we use the standard BERT model to predict a sequence of BIO tokens. We tokenise a given sentence using the  \texttt{TreebankWordTokenizer} from the \texttt{nltk} package available for the Python programming language. For the ICL setup we force the T0 model to consider only the words in the given sentence by tokenizing the sentence and feeding it into \texttt{force\_words\_ids}, additionally we also force the decoding step to not include stop words in addition to special characters like $"$ that appear in the sentence.

During decoding, we set the temperature $\tau=0.7$, top\_p=0.9, number of beams equal to 5 to provide a variety of sentences following the same narrative.

\paragraph{Stance prediction}
For the Stance prediction we restrict decoding to one word only, and giving the model two choices \textit{for} or \textit{against} for the T0 model. For the baseline model we simply attach a LM head and do binary classification $0=$\textit{for} and $1=$\textit{against}.

\paragraph{Narrative prediction}
During finetuning we switch the standard T0 model out with T0-few with the LoRA setup and mainly keep the defaulthyperparameters but reduce the batch size to 4 and train a model for each topic for for 10 epochs. Each model takes around 3hrs to train on the 10k training sen-tences. In addition to this setup we also include sentences that we generated sentences using the topic, predicted stance and aspect using the CTRL-UKP model, T0-3B, T5-flan-3B and 175B Bloom model. The tweets we predict the stance and aspect is from the test set. Using these attributes we can generate similar sentences to the teat set to help enhance performance. We simply copy the target narratives as labels for the generated sentences and include them in the training dataset.

To give an example of the runtime for our code it takes 12 hours to complete 10 epochs for the T0-3B model using 1 TitanRTX-24GB and 1 Xeon E5-2620 v4 8c/16t - 2.1 GHz CPU, and 8 hrs  using 1 A100-40GB and the same CPU for the T0-11B parameter model.
We always have access to a minimum of 48GB of RAM but run our experiments using 64GB RAM.

\begin{table}
\tablestyle{6pt}{1.02}
\scriptsize
\begin{tabular}{cc}
config & value \\
\shline
optimizer                    & AdamW   \\
base learning rate           & 3e-4                                  \\
weight decay                 & 0.001                                    \\
optimizer momentum           & $\beta_1, \beta_2{=}0.98, 0.999$          \\
adam epsilon                 & 1e-8 \\
batch size                   & 8                                    \\
learning rate schedule       & linear decay                            \\
warmup steps percentage      & 10\%                                    \\
Number of epochs             & 10                           \\
batch size                   & 8                                      \\
maximum sequence length      & 128                                     \\
maximum gradient norm        & 1                                      \\
LoRA rank                    & 4                                   \\
LoRA init scale              & 0.01                                   \\
LoRA layers                  & Self/Enc-Dec attention layers                  \\
LoRA scaling rank            & 1\\
\end{tabular}
\vspace{-.5em}
\caption{\textbf{Fine-tuning setting.}}
\label{tab:impl_finetune} \vspace{-.5em}
\end{table}

\section{Search Query and Narrative Synonyms}

\begin{table}[h]
    \centering
    \adjustbox{width=\columnwidth}{
        \setlength\tabcolsep{2.5pt}
        \begin{tabular}{ll}
        \toprule 
        \bf Topic & \bf Search query \\
        \addlinespace[0.1cm]
        AGI & \\
        &AGI replace humans, AI replace humans, AGI technology \\
        & AGI threat,AI threat,AGI beyond human intelligence,AGI rule the world\\
        &AI useful,AI better than people,society help AI
        ,AI beats human\\
        &human better AI, AGI achieve human
        ,AGI myth, human ethics AGI\\
        &AI threat humanity, AI solves problem
        ,AI help climate, AI hype bad\\
        &AI hype up, AI hype good, AGI future good, AI no common sense \\
        &AI trust bad, AI superhuman, AI wants things that are absurd to humans\\
        & AI billionaire control, AI just tool, AI wealth concentration\\
        &AI wealth inequality, automation wealth inequality ,ai if statements \\
        &ai uncontrollable, AI make me laugh, AI art steal \\
        &AI mashup, AI demotivate, AI problems fix, AI fix our problems \\
        &AI human nuance, AI data new oil, AI coming for job \\
        \midrule 
        Alternative meat & \\
        & stop subsidizing meat,alternative meat fake, alternative meat unhealthy\\
        & meat is murder, soy meat replacement, reduce meat consumption climate \\
        &meat no sustainable, meat is unhealthy, food pyramid scheme \\
        & subsidize green nutrition, increase production of meat \\
        & exempt meat production from carbon taxes, carbon tax to food production \\
        &invest Meat alternatives, Meat alternatives subsidized \\
        &Plant based food subsidized,introduce meatless mondays \\
        &Vegetarian vegan food encouraged,discourage vegan diet, subsidize fruits vegetables \\
        &meat overconsumption, Plant based food encourage, Meat alternatives encourage \\
        &plant based food sustainable, plant food is great, fresh organic food is good\\
        &meat alternative food is good,red meat is bad, animals are not ingredients \\
        &eat healthy food, raw food diet,flexitarian meat alternative \\
        &big pharma alternative meat,alternative meat forced \\
        &plant based food processed, plant based food remove meat \\
        &animals eat meat humans too, eat plant save planet,eat \\
        &meat save plant, meat ruining planet,alternative meat bugs \\
        \midrule
        Attractiveness & \\
        &spotlight effect, male gaze, male gaze exploiting, female gaze \\
        &males observing attractive female, attractive hypermasculinity \\
        &attractive masculinity,female sexual object, beauty standard money \\
        &beauty standard protection, beauty standard shoe,
        beauty standard cloth \\
        &beauty standard events, beauty standard desire,
        beauty standard objectify\\
        &beauty standard stress,beauty standard stable, beauty standard fake\\
        &beauty standard safe,sexual objectification patriarchy \\
        &beauty standard disrespect,beauty standard gender role \\
        &beauty standard escape, beauty standard equality, beauty standard dominance \\
        &beauty standard ownership,beauty standard media, beauty standard unrealistic \\
        &beauty standard transphobic, beauty standard harassed \\
        &beauty standard academia,beauty standard university \\
        &beauty standard cheating, beauty standard fetish,
        hygiene no beauty standard \\
        &Toxic Masculinity attractive,attractive Bechdel test,enforcing stereotyp beauty \\
        \midrule
        \bottomrule
        \end{tabular}
    }
    \caption{Twitter search keywords} \label{tab:twitter-search-query-1}
\vspace{-3mm}
\end{table}

\begin{table}[h]
    \centering
    \adjustbox{width=\columnwidth}{
        \setlength\tabcolsep{2.5pt}
        \begin{tabular}{ll}
        \toprule 
        \bf Topic & \bf Search query \\
        \addlinespace[0.1cm]
        corporate culture & \\
        &corporate culture HR,company culture HR,work culture HR\\
        &corporate culture toxic,company culture toxic,work culture toxic\\
        &corporate culture unlawful,company culture unlawful,work culture unlawful\\
        &company culture risk,corporate culture risk,work culture risk\\
        &corporate culture speak up,company culture speak up,work culture speak up\\
        &corporate culture abuse,company culture abuse,work culture abuse\\
        &anti union company,don't trust non profit,corporate culture no trust\\
        &company culture no trust,work culture no trust\\
        &corporate culture greed,company culture greed\\
        &work culture greed,corporate culture millenialcompany culture millenial\\
        &work culture millenial,their company do what they want\\
        &corporate culture manager,company culture manager,work culture manager\\
        &corporate culture stress,company culture stress,work culture stress\\
        &corporate culture hard work,work pregnant,side hussle culture\\
        &work culture loyalty,corporate culture loyalty,company culture loyalty\\
        &work culture remote,corporate culture remote,company culture remote\\
        &work culture ethic,corporate culture ethic,company culture ethic\\
        &work culture family,corporate culture family,company culture family\\
        &corporate culture cult,company culture cult,work culture cult\\
        &corporate culture fun,work culture fun,company culture fun\\
        &work culture perks,company culture perks,corporate culture perks\\
        &job hop look bad,company dress code,corporate mass firing, quite quitting\\
        &let it rot job,corporate culture disgusthing\\
        &work culture disgusthing, company culture disgusthing\\
        \midrule
        crypto & \\
        & china crypto ban,china crypto mining,el salvador crypto legal\\
        &crypto steal constitution,crypto banking the unbanked\\
        &crypto financially free,crypto diversify asset,crypto people of color\\
        &crypto trust technology not people,crypto access financial\\
        &crypto bank failure,crypto better digital payment,crypto wealth builder\\
        &crypto upwards mobility,crypto is an investment,crypto digital gold\\
        &crypto short the bankers,crypto not democracy,crypto ruthless investors\\
        &Bitcoin is a Platypus,crypto should be regulated,crypto needs rules\\
        &crypto is a scam,crypto is for terrorists,crypto is for criminals\\
        &crypto rich bailouts,crypto stock bubble,crypto unsustainable environment\\
        &crypto ponzi scheme,crypto pump dump,crypto influencer ponzi\\
        &crypto carbon tax,crypto great reset ,crypto own nothing happy\\
        &crypto money laundering,crypto funding party\\
        &crypto same as database, crypto is toxic\\
        \midrule
        baby formula & \\
        &baby formula scam,baby formula poison,baby formula breat milk\\
        &baby formula inflammatory,baby formula infection,baby formula virus\\
        &baby formula sustainable,baby formula weight\\
        &baby formula replacement feeding,baby formula economic, baby formula hospital\\
        &baby formula shame,baby formula guilt,baby formula husband feed\\
        &formula feeding mental health,breastfeeding mastitis,breast milk propaganda \\
        &breast milk infection,breast milk risk ,breast milk health\\
        &breast milk is best,breast milk germ,baby formula propaganda\\
        &breastfeeding guilt,breastfeeding negativity, breastfeeding anxiety\\
        &breastfeeding public,breastfeed good citizen,breastfeeding shame \\
        &breastfeeding sleeping,breastfeeding formula all nothing \\
        &breastfeeding gender role,politically correct breastfeeding \\
        \bottomrule
        \end{tabular}
    }
    \caption{Twitter search keywords} \label{tab:twitter-search-query-2}
\vspace{-3mm}
\end{table}

\begin{table}[h]
    \centering
    \adjustbox{width=\columnwidth}{
        \setlength\tabcolsep{2.5pt}
        \begin{tabular}{ll}
        \toprule 
        \bf Topic & \bf Search query \\
        \addlinespace[0.1cm]
        Influencer & \\
        &Influencer real job, Content creator full-time job \\ &Influencer popular career choice, Career social media influencer \\
        &Social media influencer pay, Social media influencer doesn’t pay well\\
        &social media influencers real job, Social media influencers unemployed \\
        &Job title influencer, quitting job influencer\\
        &Influencer marketing is big money, Influencer marketing not authentic\\
        &Social media influencer cute name employed, Self-employed influencer \\
        &Social media youths employed, adults influencer jobs\\
        &followers get a job,quitting jobs influencer jobs \\
        &social media influencers jobs money, influencer deal hate\\
        &social media job followers, social media influencers work hard \\
        &hard work influencer, Influencer new job career \\
        &popular career social media influencer, Influencer full time\\
        &influencer job pays, Influencer boring job\\
        &social media influencers getting paid, influencer easy money\\
        &influencer no respect, social media influencer celebrity\\
        &social media influencer waste of time \\
        \bottomrule
        Mental health &\\
        &Mental health-related sports ,checking mental health athletics \\
        &Mental health for athletes important,Mental health concern athletes \\
        &Mental health concern for student athletes\\
        &Sport reduces stress depression, sports affect mental health\\
        &sports healthy mind,Athletic mental health awareness\\
        &recognize mental health athletic, Prioritize mental health sports schools\\
        &sports, mental health ,Sports coach mental health\\
        &Initiative mental health sports,Well being athletic \\
        &Mental health identify issue sports, Sports support mental health \\
        &Sports stigma mental health, Stigma mental health atheletics\\
        &University atheltics metal health awareness \\
        &Stigma challenges sports mental well being\\
        &male dominated sport toxic, vulnerability  weakness sport\\
        &athletes no real problems, athletes trans problems\\
        &sports mental health flu ,sports mental health kill\\
        &sports mental health of money,Sport mental health brutal\\
        &Sport drug mental health, Sport racism mental health\\
        &athlete burnout young age,athlete burnout young\\
        &athlete blame media,athlete work late\\
        &sport alienation,Sports mental health religion\\
        &Sport mental health religion\\
        \midrule
        Transportation & \\
        &public transportation good,public transportation work\\
        &cheap public transportation, comfortable public transportation \\
        &bus better than car, public transportation environment\\
        &buses safer driving,trains better than flight\\
        &train better climate,Climate Action Public Transport\\
        &public transport safer, car culture climate, public transit affordable\\
        &flights less time trains, trains more expensive\\
        &buses carry more people,cars carry less people\\
        &cycling decrease car traffic, cycling better air quality \\
        &public transport less pollution, public transport less CO2\\
        &public transportation personal space, public transportation germs \\
        &public transportation disease, public transportation covid \\
        &public transportation rural, buses middle class\\
        &buses poor people,cars rich people ,car only wealthy\\
        &less drivers safer streets,public transportation night unsafe\\
        &public transportation night comfortable, tax car poor\\
        &use bike dangerous,public transit not profitable \\
        &public transport profitable ,highways profitable\\
        &car centric bad, public transportation useless\\
        &car give freedom independence, car give independence\\
        \end{tabular}
    }
    \caption{Twitter search keywords} \label{tab:twitter-search-query-3}
\vspace{-3mm}
\end{table}

Table \ref{tab:twitter-search-query-1} - \ref{tab:twitter-search-query-3} lists the Twitter queries we used to retrieve the initial training data.
Note that many of the words are either in their stem or shorted format in order to ensure a wider range of search results being returned. Per default twitter filter out sentences that does not contain tokens from the query. 

\section{Annotation Details}
For our crowdsourcing of narrative annotations and human evaluation we use Amazon Mechanical Turk .Workers had to take a qualification test, have an acceptance rate of at least 80\% based on 5 question, be located within the US, have successfully completed more than 1000 HITs before and have an approval rate of 98\%. 
We paid 1 dollar per HIT for the dataset task which is to classify one tweet into one in roughly 40 narrative categories. Initially time spend on a HIT is much higher than when they complete their 25th hit as workers learn to memories the categories. For the human 
evaluation we get annotations from two  crowdworkers and pay 2 dollars per HIT. Consent was obtained from the crowdworkers by including the warning for the pretest annotation: "By completing this test you will agree that subsequent HITs using this pretest as a prerequisite can be used for data collection in relation to research projects", similarly for we get consent from people whose data we are using though the Twitter Term of Services. The data collection procedure was approved by our internal ethics review board.

During our annotation of the narrative labels we discovered that the returned answers tend to be biased towards the top 10 first possible answers that could be selected in the HIT. To mitigate potential bias we manually went though the top 3 most frequent answer for each topic in the validation set and relabel the corresponding tweets.

\section{Crowdsourced Annotations}

For this paper, gathering annotations has happened over three annotations rounds, each focusing on different sections of the paper.

\subsection{Pretest}
The first crowdsourcing task is that of a pretest, which is used to determine if workers are suitable for our main annotation mask. It is based on data from \cite{hansen-hershcovich-2022-dataset} and focuses on correctly classifying two different types of labels: Pro/con and evidence.

\subsubsection{Pro/Con}
Pro/con is a binary label. The tweet is annotated as (+1) for pro when a clear claim has a positive or supportive stance towards the topic. It is annotated as (-1) when it has a clearly antagonistic or attacking stance towards it the topic. 
We exclude data for which has no clear stance.

\paragraph{Instructions for annotators:} 
Given a tweet your task is to annotate it with its stance in relation to the topic {topic}. The stance is either pro or con (for or against a topic). In this case select pro if you find that the tweet is supportive towards the topic, and con if it is hostile instead. Remember that a tweet with hostile remarks can still be supportive of the topic, as we want to find the stance towards the topic and not the tweet itself. 

\subsubsection{Evidence}
Evidence as a label has 6 classes. The tweet is annotated using any of the labels:  Normative, Study, Expert, Fact, Anecdotal or unrelated/no evidence. 
The description for each of these labels are taken from \cite{hansen-hershcovich-2022-dataset}:
Anecdotal refers to "a description of an episode(s), centred on individual(s) or clearly located in place and/or in time."
Expert refers to a "testimony by a person, group, committee, organisation with some known expertise / authority on the topic.
Study refers to "results of a quantitative analysis of data, given as numbers, or as conclusions"
Fact refers to "A known piece of information about the world without a clear source for the information"
Normative refers to "an added description for a belief about the world"
No evidence refers to "the tweet does contain evidence, but it is not related to the topic, or it does not have any evidence."

\paragraph{Instructions for annotators:} 
The task is to annotate a tweet with the type of evidence it contains. 
Evidence is a statement used to support or attack a topic or claim. Evidence can be present in combination with a claim, or it can also be self-contained if it is just stating facts or referencing studies related to the topic. 
If the evidence is unrelated to the discussed topic, it is marked as unrelated. 
If you feel that multiple types of evidence is present in the tweet, choose the one that you think best describes the main piece of evidence in the tweet.
Remember that your task is to annotate the type of evidence that is in the tweet regardless of your views and if the evidence is true or not.

\subsection{Narrative annotation}
The main crowdsourcing task of this paper is essentially claim classification.
Given a tweet the workers determine if the tweet is a claim or argument with an evidence type that is not a study (as taken from the definition of study in \cite{hansen-hershcovich-2022-dataset}). Then if the tweet is a claim then they should select the most similar claim from a list of options. If no option is suitable, they should select "No claim in list is similar to the tweet".

\paragraph{Instructions for annotators:} 
The task here is to annotate a tweet given a list of claims that the tweet might be similar to. Of course, each tweet can be relevant for more than one claim, but it can also be irrelevant and should be annotated as such.
Therefore, given that the topic is \{\} select the claim which you find the tweet most similar to (regardless of your views on the list of claims, the topic and the tweet itself). Remember that the surrounding context of a tweet can be missing, and that people may be sarcastic.

\subsection{Human evaluation of generated claims}
The last crowd sourcing campaign is the human evaluation in which we evaluate how well a generated claim compared against the original claim (It is generated from the predicted stance and aspect from the original claim ). 
We follow primarily \cite{skitalinskaya-etal-2022-optimization} for definition of argument quality, meaning and fluency, but also
\cite{schiller-etal-2021-aspect} for fluency and persuasiveness.
These generated claims are then used for finetuning a LLM for improved narrative prediction.

\paragraph{Instructions for annotators:} 
In this task, you will identify if a generated claim is similar to or has improved, without changing the overall meaning of the text.
Each field contains a par of tweets, one being the original and the other a synthetic tweet that is trying to mimic it.
Please rate each candidate along the following four perspectives: argument quality, fluency, meaning and persuasiveness. 

Argument Quality has a scale from 1 to 5: 1 (notably worse than original), 2 (slightly worse), 3 (same as original), 4 (slightly improved), 5 (notably improved)
Does the generated claim improve over the original claim?
Things to look for include: specifying a fact, simplifying the sentence, adding clarity, adding additional information such as facts, adding, editing or removing links for external resources.

Meaning has a scale from 1 to 5: 1 (entirely different), 2 (substantial differences), 3 (moderate differences), 4 (minor differences), 5 (identical)
Here we wish to measure if the generated claim  have the same overall meaning as the original. Adding extra information that does change the objects or events described in the claim should not penalise the score.

Persuasiveness runs from 1 to 3.  1 (generated text less persuasive than original), 2 (equally persuasive), 3 (generated text is more persuasive) (choose one argument as being more persuasive or both as being equally persuasive.)
Here we wish to measure if the generated claim is more useful in a debate about a certain topic than the original claim. Adding additional text that explains an event or fact more in depth should be rewarded. 

Fluency runs from a scale form 1 to 3: 1 (major errors, disfluent), 2 (minor errors), 3 (fluent)
Here we want you to to compare the generated sentence with the original one and ask if the sentence is written in fluent English and makes sense? You should consider rewarding the generated claim in case of improved grammar, spelling and punctuation of generated claim over the original claim.

\section{Narratives per topic}
\begin{table}[h]
    \centering
    \adjustbox{width=\columnwidth}{
        \setlength\tabcolsep{2.5pt}
        \begin{tabular}{ll}
        \toprule 
        \bf Topic & \bf Narrative \\
        \addlinespace[0.1cm]
        Abortion & \\
        &Abortion reduces crime\\
        &Abortion should not be allowed\\
        &Everyone has a right to life\\
        &A fetus is a real persons\\
        &Abortion is painful for the fetus\\
        &Abortion is not murder\\
        &Abortion reduces the value of human life\\
        &Women that go through abortion face social stigma or guilt\\
        &Supporting abortion is societal pressure\\
        &Abortion gives mothers the option of giving birth to healthy children\\
        &No abortion option for poor women is injustice\\
        &A fetus is not a real persons\\
        &Abortion is murder\\
        &Planned children lead better lives\\
        &Modern medicine makes abortion is less of a risk\\
        &Women choose what to do with their bodies\\
        &Couples that cant get kids want to adopt \\
        &Do not have kids if you fear they will be born with defects\\
        &Fathers have no say if the mother wants abortion\\
        &Abortion is not painful for the fetus\\
        &Removing abortion can put some pregnant woman at risk\\
        &Women that abort have no dignity\\
        &abortion leads to mental diseases\\
        &Abortion encourages more sex\\
        &Authority are against performing abortion\\
        &anti-abortion is counterproductive\\
        &Abortion is inhumane\\
        &restricting abortion enforce traditional gender stereotypes\\
        &women that have been raped should have right to abort\\
        &Abort is morally wrong\\
        &abortionists are in it for the money\\
        &Fetuses should be protected\\
        &Children who almost got aborted might feel rejected\\
        &pro-life views makes no sense\\
        &Parents must know if their child has an abortion\\
        & No claim in the list is describing the tweet\\
        \midrule
        AGI & \\
        &AI is just hype\\
        &AI art unlike human art does not have any value\\
        &AI is just if else statements\\
        &AI has no common sense\\
        &Current AI is not superhuman\\
        &AIs do not have empathy\\
        &AI is bad because it is not as good as human\\
        &AI can make you laugh\\
        &AI will not replace humans but augment them\\
        &AI is bad as it replaces artist\\
        &AGI is just a myth\\
        &AI is for the most part uncontrollable\\
        &AI will create more problems than it solves\\
        &You cannot trust AI\\
        &AI will not take your job\\
        &Data is important to make good AI\\
        &AGI will rule the world\\
        &We will get AGI sooner than excepted\\
        &AI is a threat to humans\\
        &AI will fix our problems\\
        &AI cannot recreate human nuances\\
        &AI will take your job\\
        &AI will demotivate you from working\\
        &AI is stealing from artist\\
        &You cannot trust people who hype up AI\\
        &AI will help us solve climate change\\
        &AI will live up to its hype\\
        &AI is already superhuman\\
        &AI is just a tool\\
        &AI is power hungry just like the billionaires who control it\\
        &AI furthering the wealth inequality\\
        &AI is a general purpose technology like electricity\\
        & No claim in the list is describing the tweet\\
        \bottomrule
        \end{tabular}
    }
    \caption{First list of narratives}
\vspace{-3mm}
\end{table}

\begin{table}[h]
    \centering
    \adjustbox{width=\columnwidth}{
        \setlength\tabcolsep{2.5pt}
        \begin{tabular}{ll}
        \toprule 
        \bf Topic & \bf Narrative \\
        \addlinespace[0.1cm]
        Alternative meat & \\
        &we could stop subsidising highly processed foods\\
        &Investing in Meat alternatives is good and profitable\\
        &meat production is not sustainable\\
        &alternative meat is not viable for a healthy diet\\
        &big pharma is behind the alternative meat\\
        &animals eat meat so humans should too\\
        &Eating meat is immoral\\
        &alternative meat does have enough proteins\\
        &plant based food is made to remove meat\\
        &meat cause cancer and can be deadly\\
        &plant based food are sustainable food\\
        &red meat is bad\\
        &We should subsidise Meat alternatives and Plant based food\\
        &Being Vegetarian or vegan allows you to be healthy\\
        &plant food and meat alternatives is great\\
        &animals are not ingredients\\
        &You do not need meat to hit the gym often\\
        &transport of goods is more harmful to the planet than meat or plants\\
        &alternative meat is a pyramid scheme\\
        &alternative meat tastes bad\\
        &alternative meat is unhealthy as a diet\\
        &alternative meat is forced upon the consumer\\
        &Plant based meals are highly processed and is not good\\
        &We should eat plants to save planet\\
        &alternative meat is fake\\
        &Eating fewer plants and more meat will save the plant\\
        &We should reduce meat consumption to protect the planet\\
        &We should import a carbon tax to food production\\
        &We should increase production of meat\\
        &We should stop subsidising meat to allow for alternative meat\\
        &Eating meat is murder\\
        &exempt meat production from carbon taxes\\
        &Being flexitarian allows you to get enough nutritions\\
        &fresh organic food is good\\
        &A vegan diet is unsustainable for the planet\\
        &soy meat will not and cannot replace meat\\
        &Eating bugs instead of meat will never be reality\\
        & No claim in the list is describing the tweet\\
        \midrule
        Attractiveness & \\        
        &Women and especially lesbians are exploiting the male gaze\\
        &Forcing your standard of beauty on every women is trans phobic\\
        &The male gaze encourages physical and sexual violence against women\\
        &the lack of beauty standards warrants cheating\\
        &beauty standards are pathetic and fake\\
        &Academia is like any other industry where beauty standards play into how women are treated\\
        &the female gaze and male gaze are distorted terms used on social media\\
        &Just because people do not fit the beauty standard, does not mean that you can disrespect them.\\
        &misogynists hate women that take back ownership of their bodies and reject beauty standards\\
        &beauty standards serve to perpetuate a misogynistic society\\
        &Hygienic actions like shaving is beyond beauty standards or gender roles\\
        &Beauty standards are sexist\\
        &We should be free from sexual objectification and beauty standards\\
        &Corporations try to make you buy stuff though beauty standards\\
        &beauty standards are unrealistic\\
        &feminism and gender bending is enforcing stereotypical beauty standards\\
        &beauty standards are toxic\\
        &women who do not meet conventional beauty standards are not women\\
        &beauty standards is nothing but a money making scheme\\
        &beauty standards that cater to minorities are trained to be inclusive\\
        &Women who don’t fit societal beauty standards get catcalled and harassed\\
        &social media attempts to hierarchize beauty to maintain dominance over others\\
        &beauty standards are hard to escape\\
        &Masculinity is not toxic but attractive\\
        &Beauty standards are bad and stressful for young people\\
        &No natural humans look like that\\
        &Beauty is everywhere\\
        &Beauty standards are racist\\
        &Hypermasculinity in of itself is the problem\\
        & No claim in the list is describing the tweet\\
        \midrule
        Cloning & \\
        &clones can perfect can give humans preferable qualities\\
        &Cloning can save lives or cure humans\\
        &Scientist that clone are acting unethically\\
        &Couples without kids would rather use cloning than employ surrogates or IVF\\
        &transplanting organs can be made easier and more successful by cloning\\
        &Cloning can potentially create premature ageing\\
        &Cloning in morally wrong\\
        &cloning could provide childless couples with an enhanced or enlarged family\\
        &cloning can be use to reduce risks\\
        &Cloning will cause parents to customise their children\\
        &Cloning is medicine; advances in cloning are advances in medicine\\
        &People could keep on living due to cloning\\
        &Cloning affects negatively to the reproductive processes\\
        &Better cloning techniques offer higher chances of success with less moral hazards\\
        &People that get cloned maintain their personality\\
        &Cloning is playing God\\
        &Cloning is for evil purposes\\
        &Humans will become a product with cloning\\
        &Cloning animals is cruel\\
        &Cloning someone is not a safe thing to do\\
        &Cloning is a natural\\
        &Cloned people do not have souls\\
        &Cloning will be accepted some day\\
        &Cloning is akin to murder or manslaughter\\
        &You lose a sense of individuality when cloned\\
        & No claim in the list is describing the tweet\\
        \bottomrule
        \end{tabular}
    }
    \caption{Second list of narratives}
\vspace{-3mm}
\end{table}

\begin{table}[h]
    \centering
    \adjustbox{width=\columnwidth}{
        \setlength\tabcolsep{2.5pt}
        \begin{tabular}{ll}
        \toprule 
        \bf Topic & \bf Narrative \\
        \addlinespace[0.1cm]
        Corporate culture & \\
        &non profit companies and for profit companies are equally untrustworthy\\
        &Remote work makes it difficult to maintain company culture \\
        &Jobs needs to be treated with respect\\
        &Companies simply want drones that do not ask questions\\
        &there is a lot of hustle culture for millennials\\
        &Working remove does not mean that company culture are not important\\
        &If you want a great company culture you should hire great people with a good work ethic\\
        &Corporate culture is bad\\
        &corporate culture is racist\\
        &Companies can do what they want to do\\
        &millennials do not want to work\\
        &side effect of hustle culture is often a counterproductive narrowing of focus\\
        &corporations are getting tax cuts while you are getting fired\\
        &So many popular companies actually have a terrible culture and a bad work ethic\\
        &Victims of abuse are ignored and silenced\\
        &corporations that mass fire employees say that Nobody wants to work in the media or get record high profits\\
        &mass firings are commonplace in corporate takeover\\
        &Our company culture is good because we have fun \\
        &corporate culture do not reward hard work\\
        &job hopping looks bad on your resume\\
        &Companies are anti union\\
        &corporate culture so rotten to the core by greed\\
        &Getting a stable job is getting harder over time\\
        &A fun company culture does not care about your work life balance\\
        &Some office cultures are like a cult and are not healthy for you\\
        &loyalty to the company trumps everything\\
        &Managers of a company cause nothing but trouble\\
        &perks from a company are useless\\
        &Young people refuse to enter or stay in the workforce\\
        &Companies that make you feel like a family is good\\
        &Companies that say they that you are a family is brainwashing you to comply\\
        &Hard work gets you everywhere\\
        &corporate culture promising generous pay and perks in order to mask workers disposability and exploitation\\
        &loyalty to the company means absolutely nothing in this day and age\\
        &You can make a remote workers feel part of the team without being in the same physical space\\
        &An employee is a representative of a company and should respect the dress code and look professional\\
        &work dress code is discriminatory\\
        &Corporations that do mass firings are greedy\\
        &Do not trust companies\\
        & No claim in the list is describing the tweet\\
        \midrule
        Crypto & \\
        &People will sell you out but you can trust crypto\\
        &crypto is used solely for pump and dump schemes\\
        &crypto allows a complete transformation of the global economy though the great reset\\
        &Influencer are scamming their fans using crypto\\
        &crypto is ponzi scheme\\
        &crypto is a scam\\
        &crypto is for money laundering\\
        &crypto is undiscriminatory to people of color\\
        &crypto should be regulated\\
        &crypto is legal way to pay\\
        &crypto is hijacked by ruthless investors\\
        &crypto is for terrorists\\
        &crypto is toxic\\
        &crypto is just a way to diversify your assets \\
        &crypto can bank the unbanked\\
        &crypto is for criminals\\
        &crypto is a mean for the rich to get bailouts\\
        &crypto is unsustainable for the environment\\
        &crypto better digital payment than credit cards\\
        &crypto is used to fund political parties\\
        &crypto is yet another tech bubble ready to burst and fail\\
        &Bitcoin is like a Platypus it is not real\\
        &crypto is simply digital gold\\
        &crypto allows one to become financially free \\
        &crypto allows easy access to financial services\\
        &crypto can help fighting greedy bankers\\
        &crypto is an investment\\
        &crypto is used to steal the constitution\\
        &Crypto is not democratic\\
        &crypto should be carbon taxed\\
        &crypto is useful when banks are failing\\
        &crypto mining should be banned\\
        &blockchain is no different from a database\\
        &crypto is a wealth builder\\
        &WEF want you to own nothing and be happy, crypto avoids this problem\\
        &crypto provides upwards mobility\\
        & No claim in the list is describing the tweet\\
        \bottomrule
        \end{tabular}
    }
    \caption{Third list of narratives}
\vspace{-3mm}
\end{table}

\begin{table}[h]
    \centering
    \adjustbox{width=\columnwidth}{
        \setlength\tabcolsep{2.5pt}
        \begin{tabular}{ll}
        \toprule 
        \bf Topic & \bf Narrative \\
        \addlinespace[0.1cm]
        (Baby) Formula & \\
        &breastfeeding is natural and is not politically correct\\
        &women are pressured into breastfeeding and gets stressed\\
        &baby formula is costly\\
        &baby formula is killing babies\\
        &It does not have to be all or nothing \\
        &the political right vote against bills to make things more expensive\\
        &breastfeeding and baby formula is risky\\
        &baby formula ends up in foreign countries instead of the US where it should be\\
        &Some people are allergic to breast milk and need formula\\
        &People hate babies when they make abortion illegal and remove baby formula\\
        &baby food industry is promoting propaganda\\
        &breastfeeding can cause HIV\\
        &baby formula is poisonous\\
        &baby formula is good as fathers can feed their baby\\
        &baby formula is best if you cannot breastfeed\\
        &It is important to secure enough baby formula in an economic crisis\\
        &breast milk is best\\
        &The political left is out to remove babies\\
        &People are reselling baby formula to other countries for higher prices\\
        &Do what is best for you and the baby\\
        &The baby formula shortage is one big scam\\
        &Some people cannot tolerate formula and need milk\\
        &breast milk has a lot of antibodies that can help the baby fight off infection\\
        &people publicly shame women who breastfeed in public\\
        &mental health is more important than breast feeding\\
        &breastfeeding is healthier than baby formula\\
        &baby formula can cause infection\\
        &breastfeeding will lowers risk of breast cancer\\
        &The political right have caused a baby formula shortage\\
        &breast is best campaign causes anxiety in moms who cannot breastfeed\\
        & No claim in the list is describing the tweet\\
        \midrule
        Influencers & \\
        &Influencers can be damaged by everything they say\\
        &social media in the west is like opium for kids\\
        &influencers just want to get rich quick\\
        &influencer marketing are not authentic\\
        &social media career is not sustainable\\
        &influencers are creative\\
        &influencers earn too much money\\
        &social media people are toxic and rude\\
        &an influencer is a social media celebrity\\
        &Normal jobs are boring\\
        &influencing is indeed hard work\\
        &influencers understand the use cases of products and want to help\\
        &you should not quit your job and become an influencer\\
        &social media people are just plagiarising other people\\
        &Becoming an influencer allows you to live the good life \\
        &influencers do not know hard work\\
        &influencers wants to be their own boss\\
        &dealing with hate is part of a social media job\\
        &the numbers of followers do not make you successful\\
        &influencers are not respected\\
        &influencer is not an adult job\\
        &jobless youth are spending too much time on social media platforms\\
        &If you have too few followers you should get a job\\
        &influencers are wasting their time \\
        &being an influencer is easy\\
        &influencer is not a real job title\\
        & No claim in the list is describing the tweet\\
        \midrule
        Mental Health in sports\\
        &male dominated sports are toxic for women\\
        &In sports people get into drugs when mental health declines\\
        &sport help you alleviate stress\\
        &The money made in sports should go to mental health organisations not admin staff\\
        &team sport is a brutal business\\
        &female athlete are not projected\\
        &When athletes get in trouble the blame the media\\
        &sports is  like religion it is bad for your mental health\\
        &In sports racism and mental health issues goes hand in hand\\
        &athletes do not have any problems\\
        &athletes are or must become hard workers\\
        &athletes does not have real mental health problems\\
        &Work late nights, don’t take sick days\\
        &Be tough, vulnerability is weakness\\
        &Entering sports in an early age led to burnout\\
        &getting help is stigmatising\\
        &elite athletes have unfair genetic advantages\\
        &sports athletes are manipulated\\
        &mental health is not masculine\\
        &athletes are only thriving professionally if they thrive personally\\
        &mental health should not be treated like the flu\\
        &trans people should not participate in male or female sports\\
        &be ashamed if talking about mental health\\
        &sports athletes are depressed\\
        &you do as your told as an athlete\\
        &alienation cause mental health issues in sports\\
        &athletes should not worry because they have a lot of money\\
        &talking about mental health is showing weakness\\
        &athletes that speaks up about health issues are silenced\\
        & No claim in the list is describing the tweet\\
        \bottomrule
        \end{tabular}
    }
    \caption{Fourth List of narratives}
\vspace{-3mm}
\end{table}

\begin{table}[h]
    \centering
    \adjustbox{width=\columnwidth}{
        \setlength\tabcolsep{2.5pt}
        \begin{tabular}{ll}
        \toprule 
        \bf Topic & \bf Narrative \\
        \addlinespace[0.1cm]
        Nuclear Energy & \\
        &Nuclear reactor is easy to control\\
        &Deciding what to do with regards to long term disposal of nuclear energy waste is difficult\\
        &Nuclear energy will be available for use longer than oil for example\\
        &Nuclear energy will contaminate the environment\\
        &Nuclear energy is dangerous\\
        &Nuclear energy can give us unlimited energy\\
        &nuclear energy waste can be recycled\\
        &Nuclear energy is good\\
        &Using nuclear energy to solve problems that arise is logical\\
        &Nuclear energy leads to more violence\\
        &nuclear power produce carbon free energy\\
        &nuclear energy is not efficient\\
        &nuclear power is financially burdensome\\
        &There is no significant risk with nuclear energy that cannot be said about other agents as well\\
        &nuclear energy is dirty\\
        &nuclear energy is not safe\\
        &nuclear energy makes poor nations dependant on rich nations\\
        &Every country can use nuclear energy unlike everything else\\
        &Nuclear energy relies too heavily on subsidies\\
        &There is not a good plan for storing or disposing of nuclear energy waste so we should use it\\
        &Nuclear plants only produce electricity and cannot replace oil and gas\\
        &Using nuclear power will lead to nuclear war.\\
        &renewable energy is a more viable option than nuclear energy\\
        &Nuclear energy are favoured by certain social structures like capitalism\\
        &decentralised nuclear energy production is efficient\\
        &Nuclear power is needed to stabilise climate change.\\
        &Nuclear energy should not even be considered as an energy source\\
        &Nuclear energy is much more harmful than beneficial\\
        &Green energy will make nuclear energy obsolete\\
        &Nuclear energy will increase the cancer in humans\\
        &nuclear energy is not renewable energy\\
        &Nuclear reactors are vulnerable to terrorist attack\\
        &nuclear energy is more reliable than renewable energy sources like solar\\
        & No claim in the list is describing the tweet\\
        \midrule
        Transport & \\
        &trains are better than flights\\
        &public transportation is unsustainable for rural areas\\
        &public transportation is useless car is better\\
        &cars give you the freedom of Independence\\
        &buses are safer than cars\\
        &fewer drivers equals safer streets\\
        &public transportation is comfortable\\
        &public transport results in less pollution\\
        &trains are better for the climate\\
        &public transportation has no personal space\\
        &It is important that public transit works\\
        &cars are good when there is no good alternative\\
        &flights are better than trains\\
        &public transportation is for poor people\\
        &public transportation is ridden with disease\\
        &highways are not profitable\\
        &public transit only works if affordable\\
        &using bikes are dangerous\\
        &public transportation is filled with germs\\
        &cars are worse than buses as it carries less people\\
        &buses are better than cars\\
        &public transportation is good business\\
        &taxing car and roads hurt the poor people\\
        &cars are for rich people\\
        &public transportation is unsafe at night\\
        &car centric infrastructure is bad\\
        &trains are too expensive\\
        &No good transportation is a reflection of the government\\
        &cycling will decrease car traffic\\
        &public transit is not profitable\\
        &public transportation is good\\
        & No claim in the list is describing the tweet\\
        \bottomrule
        \end{tabular}
    }
    \caption{Fifth list of narratives}
\vspace{-3mm}
\end{table}

\newpage

\end{document}